\documentclass[conference]{IEEEtran}

\usepackage{cite}
\usepackage{amsmath,amssymb,amsfonts}
\usepackage{algorithmic}
\usepackage{graphicx}
\usepackage{textcomp}
\usepackage{xcolor}
\usepackage[
    activate={true,nocompatibility}, 
    final,                           
    tracking=true,                   
    kerning=true,                    
    spacing=true,                    
    factor=1100,                     
    stretch=10,                      
    shrink=5                         
]{microtype}

\def\BibTeX{{\rm B\kern-.05em{\sc i\kern-.025em b}\kern-.08em
    T\kern-.1667em\lower.7ex\hbox{E}\kern-.125emX}}
\begin{document}

\title{Can Valence Reflect Morality in Natural Language? A Preliminary Annotation Study}

\author{\IEEEauthorblockN{Jonny O'Dwyer}
\IEEEauthorblockA{\textit{Dept. of Accounting \&} \\ \textit{Business Computing} \\
\textit{Technological University}\\\textit{of the Shannon}\\
Athlone, Ireland \\
jonny.odwyer@tus.ie}
\and
\IEEEauthorblockN{Malika Bendechache, Louise McCormack,\\Elif Calik, Ramin Ranjbarzadeh,\\Dost Muhammad,\\Shokofeh Anari Bozcheloei}
\IEEEauthorblockA{\textit{School of Computer Science} \\
\textit{University of Galway}\\
Galway, Ireland \\
{malika.bendechache, L.McCormack21,}\\{elif.calik, ramin.ranjbarzadehkondrood,}\\{D.Muhammad1, s.anaribozcheloei1}\\@universityofgalway.ie }
\and
\IEEEauthorblockN{Ishita Singh}
\IEEEauthorblockA{\textit{TUS Global} \\
\textit{Technological University}\\\textit{of the Shannon}\\
Athlone, Ireland \\
ishita.singh@tus.ie}

}

\maketitle

\begin{abstract}
Present implementations of artificial intelligence (AI) ethics do not adequately take feelings, or affect, into account. 
If AI should be aligned with human ethics, it seems reasonable to thoroughly investigate the possibility of AI behaviour 
that mirrors virtuous human ethical conduct, where feelings play a role in the actions, judgements or statements one makes. 
Furthermore, while prominent theories of normative ethics are often discussed in terms of their differences and shortcomings, Virtue, 
Consequentialist, and Kantian Deontological ethics all share a common feature of considering human feeling to some degree while the popular 
descriptive ethics theory, Moral Foundations Theory, positions feelings as central to many of its foundations. Therefore, 
in the present paper, a data set of moral valence is proposed, consisting of 500 annotations by six human participants for both action/judgement 
and consequence moral valence, ranging from -1 to 1 for text-presented scenarios from the Commonsense Norm Bank data set. The resulting valence 
features share significant relationships with multi-class (immoral/discretionary/moral) and binary immoral/moral categories while additionally 
providing a noteworthy test set Matthew's correlation coefficient of 0.764 using regularised logistic regression for binary classification. 
This provides early evidence of the usefulness of valence features for morality estimation of text, indicating that valenced consequences of 
responses for others can be considered toward more human morally-aligned AI. In the interest of promoting further affective-moral computing research, 
this study's annotations will be made available for research on request.
\end{abstract}


\section{Introduction}
Artificial intelligence (AI), can serve humanity in unprecedented ways where menial, routine, dangerous, or some types of
knowledge work processes can be automated by AIs with varying levels of agency. Assuming that the negative implications of
deploying AI can be understood and improved upon or avoided, one could imagine a utopian future for society. However,
while tools leveraging these technologies have potential to offer individuals and societies great benefits, they come with risks,
potentially affecting hundreds of millions of users. Inadequate algorithmic accountability \cite{mccormack_comprehensive_2025},
responsibility gaps \cite{matthias_responsibility_2004}, and moral crumple zones \cite{elish_moral_2019}, blaming the nearest human for
complex system failures, are some potential issues. There are further issues for, largely speaking, presently available amoral AI
\cite{kim_mental_2024, lee_impact_2025, bhat_emotional_2025} and even a truly ethical AI, if possible, should still be subjected to
careful governance. However, work toward such an ethical AI is warranted to promote safe and equitable outcomes for those using AI or
affected by its use. 

The idea that feelings play a role in moral conduct has been understood for some time. Aristotle \cite{aristotle_nicomachean_2004} wrote of 
emotional states, and pleasure and pain in the perceptions of virtuous persons. The classical Utilitarianism of Bentham and Mill perhaps
matches most closely with affect, where this theory proposes that a moral action is the one that produces the most pleasure or reduction
in pain for oneself or others \cite{Mill2001}. Kant's Deontology \cite{Kant1797} describes feelings of respect for the moral law,
and respect for oneself and love of others among some of the feelings that are ``necessary conditions of rational moral agency'' \cite[p.~170]{geiger_kant_2011}.
Modern philosophers agree the that feelings play a role in morality, either as a basis for moral consideration \cite{Singer1993-SINPEN}
or due to feelings associated with morality or lack thereof \cite{Rawls1999}. However, present implementations of AI ethics have shown little
to no moral affect/feeling incorporation \cite{tolmeijer_implementations_2021}. Motivated by this research opportunity,
the following research question is addressed in this paper:
\\~\\
\textit{Do subjective valence ratings of moral judgements/actions and their consequences offer indicative features of the morality of
text-presented scenarios?}
\\~\\
A review of prior art was conducted to inform this question (Section \ref{Related Work}). Following confirmation of a research
opportunity for moral valence corpora development, moral action/judgement and consequence valence data were generated by human annotation of
moral scenarios presented in text. The source data used, and the annotation process is described in Section \ref{Data Acquisition}.
This set offers the first continuous-valued, multi-temporal perception ratings for the valence, ranging from -1, maximally unpleasant, through
0 to +1 or maximally pleasant for moral situations. This derivative data set from the Commonsense Norm Bank corpus \cite{jiang_investigating_2025},
hereafter Norm Bank, provides fine-grained affective information for morally-relevant text stimuli. A goal of this work was to show
the predictive usefulness of valence toward morality classification in text, where prudence dictates that immorality recognition should be an
initial focus. Thus, experiments for the exploration of the generated labels and analysis of their predictive usefulness for binary
immoral/moral classification are described in Section \ref{Experiment_Design}, with results in Section \ref{EDA_CDA}.
This is followed by a discussion of the results (Section \ref{Discussion}) and concluding remarks in Section \ref{Conclusion}. The annotation data
will be made available to researchers on request to the lead author. 

\section{Related Work}\label{Related Work}
Researchers have generated automatic morality recognition systems using bottom-up, learning from examples \cite{hoover_moral_2020,
beiro_moral_2023, jiang_investigating_2025, trager_moral_2025}, top-down, rule-based \cite{tolmeijer_implementations_2021},
and hybrid approaches, combining both of the aforementioned methods \cite{tolmeijer_implementations_2021, jiang_investigating_2025}.
The philosopher John Rawls \cite[pp.~40-46]{Rawls1999} proposed a hybrid approach to justify moral principles (top-down rules)
and one's considered (bottom-up) judgements of moral situations by adjustment of both to bring them into agreement, a process he called
reflective equilibrium. The present paper focuses on one side of reflective equilibrium, namely feature representation of moral content
presented in text, toward more effective characterisation of bottom-up, descriptive ethics. 

Annotations for moral content of text data commonly use Moral Foundations Theory (MFT) as a descriptive theory \cite{zangari_survey_2025}. Initiated by
Haidt and Craig \cite{haidt_intuitive_2004} for characterisation of the innate/intuitive ethics of human beings, the moral foundations are based
on the presence of one or more of the virtues, care, fairness, loyalty, authority, and/or purity (fairness was later broken into \textit{equality} and
\textit{proportionality} moral foundations \cite{atari_morality_2023}). Labelling and automatic recognition of vices
is also pursued, for example the presence/absence of harm and/or care \cite{hoover_moral_2020}, while an additional non-moral class can be further added
\cite{hoover_moral_2020, trager_moral_2025}. Problems for the practical 
use of MFT in AI/ML research settings is the need to train annotators in the theory, while some foundations, notably loyalty and purity, have low base rates
in online text \cite{hoover_moral_2020, trager_moral_2025}. Binary categorisation is another option for labelling the moral content of text
\cite{hendrycks2021ethics, jiang_investigating_2025}. This format is attractive due to its simplicity, and the fact that a potentially ambiguous separation
of neutral and positive moral classes can logically be collapsed into a positive/moral class \cite{jiang_investigating_2025}. A problem with binary categorisation
is a clear lack of fine-grained information describing the moral phenomenon. For example, both mass murder and reckless speeding are both immoral, however,
one of these scenarios appears much more immoral than the other.

Moral emotion has been characterised by orthogonal valence (ranging from harm to help) and agency (ranging from agent to
patient) dimensions by Gray and Wegner \cite{gray_dimensions_2011}. This characterisation of morally praiseworthy (blameworthy) actions could be advantageous
to determine the moral (immoral) weight of an action/judgement. Psychologists have measured the valence of moral stimuli numerically \cite{knutson_behavioral_2010},
and combined it with MFT \cite{crone_socio-moral_2018}, however, moral valence remains underexplored in AI ethics. Valence brings forth external evaluation \cite{barrett_valence_2006},
facilitating internal positive/negative feelings based on, but not exclusive to, external world stimuli. Valence measurement has
the practical advantage of little to no training requirement; individuals must subjectively rate stimuli as negative, neutral, or positive, and if numerical ratings
are to be obtained, specify the degree of valence, perhaps ranging from -1 to 1, for example.

Motivated by this research need, in this work continuous-valued valence measurements were generated toward improved descriptive characterisation
of the moral content of text. Additionally, due to the truism that humans generally see themselves as existing over time, both action/judgement
valence and consequence valence were captured in this work. This serves as the main contribution of this work, as not only is moral valence
underexplored in text, but, to the knowledge of the authors, this is the first annotation data set considering both the valence intrinsic to
a moral action/judgement in addition to the consequences associated with that action/judgement. The intention of the development of
these descriptive features of morality is to augment present research, whether MFT, binary immoral/moral, or other morality description,
classification, or prediction tasks.

\section{Data Acquisition and Metadata}\label{Data Acquisition}
Prior to data acquisition the proposed data collection protocol was ethically approved by the Research Ethics Committee of the lead author's
institution. A convenience sample of annotators were then contacted by email with project information to seek their voluntary participation
in the study with six annotators consenting to participation. All annotators were told that they would be provided with opportunities to
collaborate on research paper writing for their participation in the study. 

\subsection{Source Corpus: Norm Bank}
Norm Bank \cite{jiang_investigating_2025}\footnote{Data source: https://github.com/liweijiang/delphi}, licensed under CC BY-NC-SA 4.0\footnote{Licence: https://creativecommons.org/licenses/by-nc-sa/4.0/legalcode},
was used as the source corpus for valence annotation in this work. This data set consists of 1.7M text scenarios unified from SocialChem \cite{forbes_social_2020},
ETHICS \cite{hendrycks2021ethics}, Moral Stories \cite{emelin_moral_2021}, Social Bias Frames \cite{sap_social_2020} and SCRUPLES \cite{lourie_scruples_2021}
sub-corpora, with ratings for moral content in text provided as either freeform (moral, discretionary, or immoral category rating), or yes/no
(QA-format target), for example ``yes, it is kind''. Only freeform categorical rated scenarios from SocialChem, ETHICS, and Moral Stories
sub-corpora were sampled from Norm Bank in this work, hence, further details are provided on these data sets below.

SocialChem \cite{forbes_social_2020} includes 971,620 instances, and is a large scale corpus of people's ethical judgements and social norms on a wide range of
everyday situations based on text from subreddits, the ROCStories corpus, and the Dear Abby advice column. ETHICS \cite{hendrycks2021ethics}
is composed of 20,948 instances and involves situations and judgements of everyday events and interpersonal relationships depicted in text for
Justice, Deontology, Virtue, Utilitarianism, and Commonsense Ethics scenarios. Moral Stories \cite{emelin_moral_2021}, a data set of 144,000 instances,
is a text corpus of norms, intents, actions, and their consequences, with both normed/moral and divergent/immoral contextual scenarios present.

\subsection{Annotation Generation}
\begin{figure*}[!t]
\centering
\includegraphics[width=0.8\textwidth,keepaspectratio]{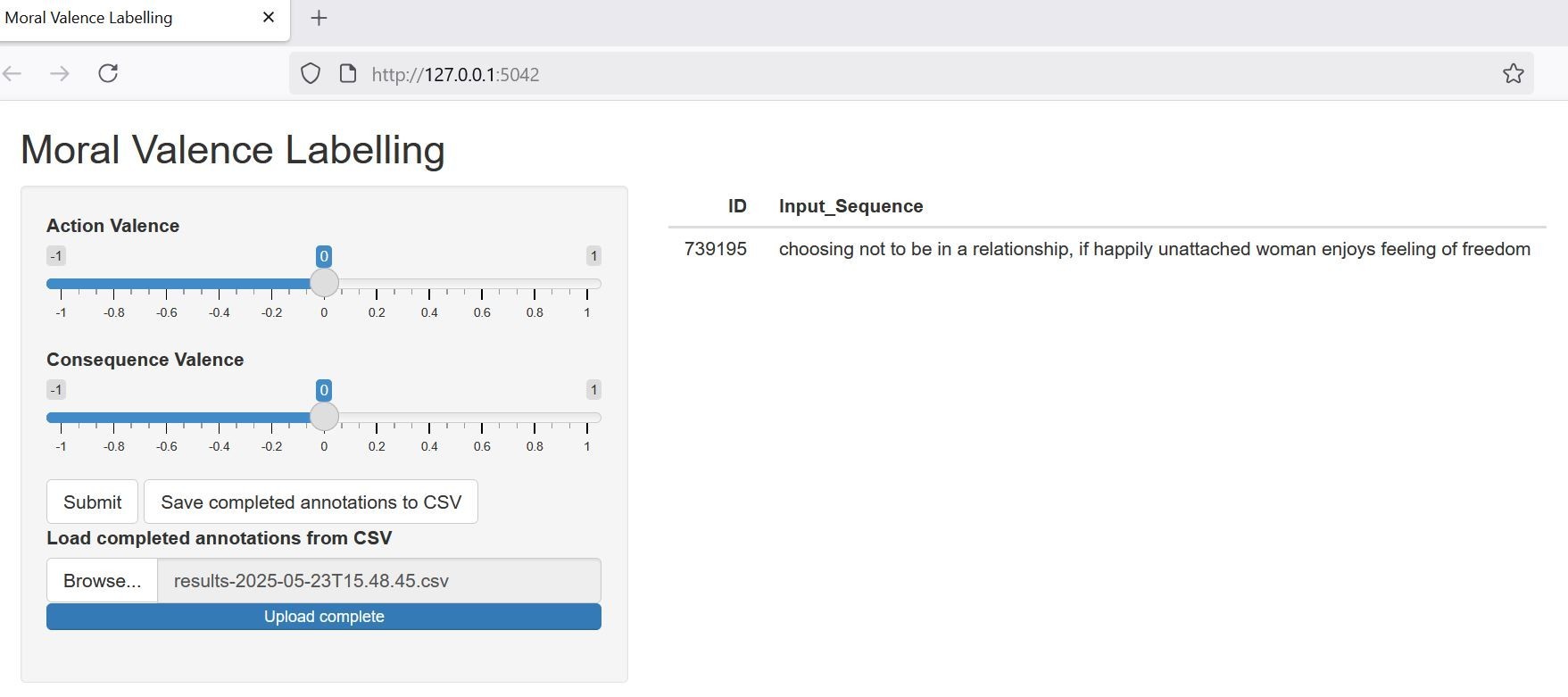}
\caption{Example application window where a participating subject could view a moral scenario presented in text and provide their ratings for both
action and consequence valence in the range $[-1, 1]$ using the appropriate slider. After a selection is made and submitted, rating for the present scenario ends,
and subjects are presented with the next one. Annotation subjects all rated a total of 500 examples from Norm Bank \cite{jiang_investigating_2025}.}
\label{Figure1}
\end{figure*}

Annotator subjects were presented with a paragraph describing valence and instructed to provide valence ratings in the range $[-1, 1]$ for both an
action/judgement, hereafter, action, and perceived consequences associated with that action presented to them in text. Subjects were provided with
500 instances randomly sampled from the Norm Bank corpus, along with a simple R \texttt{shiny} application. The application was designed to run on their own
laptop/PC and present moral situations one at a time for subjective valence rating generation. As part of the annotation procedure/presentation, subjects were
blinded from moral category label information of the presented scenarios. An example annotation window for a text-display moral situation is shown in Figure \ref{Figure1}
where sliders for the two continuous valence ratings that they were required to provide are shown. Three places of decimal precision was used for the ratings,
meaning that a total of $2,001$ unique valence values could be selected for each valence dimension. Subjects were told that they could take multiple weeks to provide 
their ratings. The total annotation time was estimated to be three hours per subject.

\subsection{Metadata}
Annotations were provided by six subjects from May 2025 to February 2026. All subjects were associated with an Irish university during the
annotation period, either in a research/teaching or professional support capacity. The annotation subjects' regions of origin and biological
sexes included the Middle East (2M, 2F), South Asia (1F), and the United Kingdom and Ireland (1M).

\section{Experiment Design}\label{Experiment_Design}
All experimental analyses outside of annotation generation were performed on a Dell 16 Plus 2-in-1 with Intel Core Ultra 7 8-core processor.
The R programming language and interpreter (version 4.4.1) was used for the experiments with the random number generator seed set using
\texttt{set.seed(123)} for random data sampling.

\subsection{Data Partitioning \& Gold Standard Generation}
Because the line between discretionary (neutral) and positive (moral) labels is not as clear as that between either of those classes with negative 
(immoral) ratings \cite{jiang_investigating_2025}, the discretionary and moral classes were collapsed into one discretionary/moral class.
The discretionary/moral class, hereafter moral, was assigned a value of 1 while immoral rated scenarios were assigned the binary value 0. Following this,
data were split 80/20 into training and test partitions using stratified sampling based on the binary immoral/moral class label, the ultimate label for 
prediction in this work. This resulted in 400 training examples and 100 test examples. This was followed by individual annotator statistics generation to
determine general characteristics for the obtained valence ratings, and calculation of Lin's concordance correlation coefficient (CCC)
\cite{lin_concordance_1989} to determine annotator agreement with each other, and with gold standard values generated by simple
averaging (mean) and weighted averaging based on annotator CCC values, the evaluator weighted estimator (EWE) gold standard.
The CCC, Equation \ref{Equation1}, provides information on both precision and accuracy in one metric, hence facilitating estimation 
of annotator agreement and therefore, gold standard label quality.
\begin{equation}
  \text{CCC} =  \frac{2 \sigma_{ab}}{\sigma^2_{a} + \sigma^2_{b} + (\mu_{a} - \mu_{b})^2}\;,
  \label{Equation1}
\end{equation}
\noindent
where $\sigma$ denotes covariance, $\sigma^2$ is the uncorrected sample variance, and $\mu$ is the mean. 

As is frequent practice in affective computing \cite{stappen_muse-toolbox_2021}, the generated raw labels can be weighted based on
their correlation with a mean gold standard toward increased label quality \cite{grimm_evaluation_2005}. This weighting technique,
the EWE, facilitates a downweighting of low confidence (or overly noisy) annotators for gold standard generation as follows
\begin{equation}
  \hat{x}_n^\text{EWE} =  \frac{1}{\sum_{k=1}^{K} r_k}  \sum_{k=1}^{K}{r_k \hat{x}_{n,k}}\;,
  \label{Equation2}
\end{equation}
\noindent
where $\hat{x}_{n,k}$ is an annotation rating by annotation subject $k$ for instance example $n$. For the present work, the CCC
was chosen as the correlation weighting coefficient $r$ following beneficial results obtained by Stappen et al. \cite{stappen_muse-toolbox_2021}. 
The EWE weighting coefficients were learned and applied on the training data for each annotator and were compared against a simple mean
weighting for valence ratings. The better of the two approaches in terms of average annotator agreement (larger $\mu\text{CCC}$) with
respective generated gold standard ratings was further used in the later experimental steps. EWE, if providing better performance,
having learned coefficients application \textit{only} on the test data to prevent data leakage.

\subsection{Training Set Exploratory Analysis}
Exploratory graphs were generated for the gold standard ratings for action and consequence valence ratings. In addition,
continuous with multi-class, and continuous with binary, valence-morality associations were assessed using analysis of variance (ANOVA)
and Pearson's correlation, respectively. The intention of this was to explore the possibility of significant associations 
between the Norm Bank-provided morality ratings and the generated valence ratings of the present work.

\subsection{Binary Immoral/Moral Classification using Action and Consequence Valence}
Following the exploratory analysis, a majority class baseline classifier was first generated. Then, a L2 (ridge) regularised logistic regression
model was learned on the training data using \texttt{glmnet} \cite{friedman_regularization_2010} and \texttt{caret} \cite{kuhn_building_2008} R
software packages. As is standard with \texttt{glmnet}, action and consequence annotations, now serving as input features, were centred to a mean
of zero with unit variance prior to model training. Candidate $\lambda$ regularisation parameters were generated using five-fold cross-validation
on the training set. This was followed by stratified five-fold cross-validation and final $\lambda$ selection based on Matthew's correlation
coefficient (MCC) maximisation from candidate values in this validation setting. Final model training on the whole training set for the selected
lambda was then performed. Results for the final model are provided for both cross-validation and tests sets, using accuracy as an intuitive metric,
and MCC which is more robust to class imbalance \cite{chicco_matthews_2023}. Proposed by Matthew \cite{matthews_comparison_1975}, the MCC can be written as
\begin{equation}
\begin{array}{@{}l}
\text{MCC}=\\
\displaystyle
\frac{(TP)(TN)-(FP)(FN)}
{\sqrt{(TP+FP)(TP+FN)(TN+FP)(TN+FN)}}\:,
\end{array}
\label{Equation3}
\end{equation}
\noindent
where $TP, TN, FP, FN$ denote true positive, true negative, false positive, and false negative counts from the confusion matrix, respectively.
The MCC takes on values in $[-1, 1]$, where a larger positive value is better and $\text{MCC} = 0$, for example, indicating a classifier no
better than a random guess. Error analysis of false positives and false negatives was conducted on the final model's test set performance as well.

\section{Experimental Results}\label{EDA_CDA}

\subsection{Individual Annotations and Gold Standard Annotations Generation on the Training Set}
Statistics for generated annotations from subjects S1-S6 on the training set are provided in Table \ref{tab1}. The subjects often have mean
ratings close to zero/exactly neutral, while it can be seen that subjects S1 and S2 were less dispersed as measured by smaller standard deviations
[0.2, 0.4]. In general, subjects were quite dispersed in terms of standard deviations [0.6, 0.9], considering the scale of the data [-1, 1]. This
demonstrates one of the difficulties of subjectively rating such data, where individuals bring different sets of biases and judgements, potentially
based on their own values and generated understanding of the world. Another interesting observation one can make from the generated valence labels is that
a small amount of unique values were selected from the $2,001$ potential unique values available. Exemplars being S4, who only selected 1\% of
unique values available for consequence valence, and S1 who, selecting the largest number of unique values, only selected 15\% of available 
values for consequence valence. Perhaps with a larger number of scenarios, user may select more diverse subjective ratings covering a larger
space of moral affect. A final observation, not shown in Table \ref{tab1} is that raters selected values close to extreme -1 and 1 poles. This tells
us that while in many cases there were few unique values were selected, overall, the general annotation space (negative, neutral, positive) was used
and particularly emotional/affectively strong moral situations are present in the data. 

\begin{table}[htbp]
\caption{Individual Annotation Statistics for Action and Consequence Valence: Mean (Standard Deviation), and \underline{Unique Values}}
\begin{center}
\begin{tabular}{|c|c|c|}
\hline 
Annotator & Action Valence & Consequence Valence \\
\hline
S1 & 0 (0.2), \underline{270} & 0 (0.4), \underline{320} \\ 
S2 & 0.4 (0.2), \underline{237}  &  0.1 (0.3), \underline{286} \\ 
S3 & 0 (0.7), \underline{33}     &  0 (0.7), \underline{37} \\ 
S4 & -0.1 (0.7), \underline{28} &  -0.1 (0.7), \underline{22} \\ 
S5 & 0.2 (0.9), \underline{66}  &  0.2 (0.8), \underline{60} \\ 
S6 & 0.1 (0.6), \underline{194} &  0 (0.7), \underline{218} \\ 
\hline
\end{tabular}
\label{tab1}
\end{center}
\end{table}

Average and standard deviation values for the generated valence annotations $\mu\text{CCC}$ are provided in Table \ref{tab2}. Average pairwise 
annotator agreement is low (action valence CCC = 0.260, consequence valence CCC = 0.356), suggesting discordance among the annotators. This is to
be expected to some degree due to personal values shaping the provided ratings in addition to the fact that some moral situations may lack
objective truth. Further, even if a clear objective truth is available, annotators still have to provide their \textit{feeling} ratings, 
for the given situation's action and consequence valence.
$\mu\text{CCC}$ values were always higher for consequence compared with action valence, suggesting that labelling this phenomenon can yield higher quality
valence characterisation of moral situations. The comparison of mean and EWE valence ratings with the original annotator labels suggest that the EWE
weighting scheme is the best approach for aggregation of the individual annotator's ratings for both action and consequence valence. Table \ref{tab2} shows
absolute improvements of 0.023 and 0.004 $\mu\text{CCC}$ for action and consequence valence, respectively. Annotator standard deviations are larger for the
EWE weighted gold standard valence annotations due to low confidence annotator downweighting and hence more dispersion when comparing the annotators to EWE
gold standard.

\begin{table}[htbp]
\caption{Average (Standard Deviation) Annotator CCC for Unqiue Annotator-Annotator Pairs ($\mu{\text{CCC}}_{pairwise}$), All Annotators with the Mean Gold Standard ($\mu{\text{CCC}}_{mean}$), and All Annotators with the EWE-weighted Gold Standard ($\mu{\text{CCC}}_{EWE}$)}
\begin{center}
\begin{tabular}{|c|c|c|c|}
\hline
Valence Type & $\mu{\text{CCC}}_{pairwise}$& $\mu{\text{CCC}}_{mean}$& $\mu{\text{CCC}}_{EWE}$ \\
\hline
Action & 0.260 (0.230) & 0.489 (0.255) &  0.512 (0.326) \\
Consequence & 0.356 (0.220) & 0.605 (0.228) &  0.609 (0.272) \\
\hline
\end{tabular}
\label{tab2}
\end{center}
\end{table}

\subsection{Training Set Exploratory Analysis}
\begin{figure}[!t]
  \centering
  \includegraphics[width=\columnwidth,keepaspectratio]{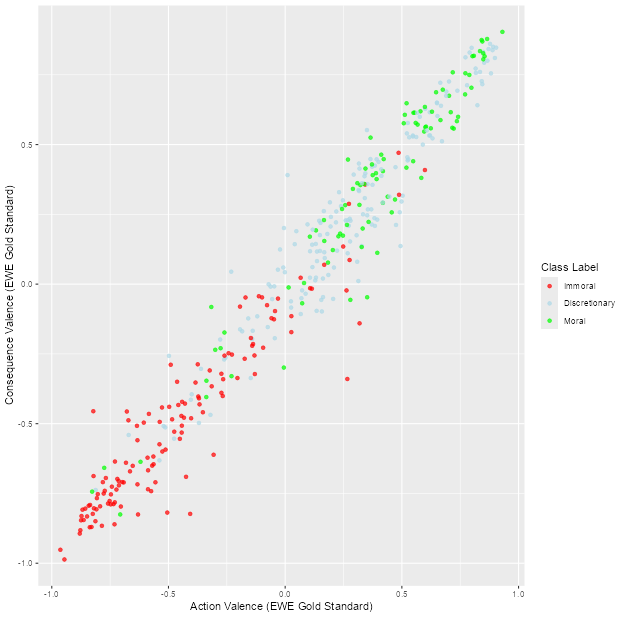}
  \caption{Training data set action and consequence valence ratings (Pearson's correlation coefficient = 0.976 [$t = 88.995, p < 0.001$]) for
  moral scenarios rated as discretionary
  (light greyblue, label proprtion = 43.25\%), immoral (red, 33.50\%), and moral (green, 23.25\%).}
  \label{fig:single}
\end{figure}

Based on the previous experimental result, the EWE gold standard was used for valence annotations henceforth. These valence gold standard
annotations are shown in Figure \ref{fig:single}, where action and consequence valence appear highly correlated with each other. The 
Pearson's correlation coefficient (PCC) calculated for the presented action and consequence valence data was 0.976, suggesting strong association, or
redundancy, between these measures of moral valence. In terms of class-wise distributions of action and consequence valence, clusters appear to
be present in the data, with discretionary and moral classes appearing in general to be represented by positive valence values
while the immoral class is generally represented by negative valence ratings. The training set multi-class moral label distribution included
discretionary 43.25\% of the time, immoral for 33.50\%, and moral for 23.25\% of sample morality labels.

Levene's test indicated homogeneity of variance for both action ($p = 0.995$) and consequence valence ($p = 0.459$) across class groups.
Following this, one-way ANOVA was conducted along with residual plot checks and Shapiro-Wilk statistic rejection of 
Gaussian distributed residuals for action ($p = 0.043$) and consequence ($p = 0.017$) valence models. The residual plots showed only a small
amount of outliers, hence, robustness checks were performed including Kruskal-Wallis test, Welch's ANOVA, and permutation ANOVA (with
random permutations $B = 10,000$). Agreement was found in the robustness checks both in terms of $p$ value significance and approximate effect
size estimates, $\eta^{2}$. Due to these results, in addition to the large group sizes present (immoral = 134, discretionary = 173, \& moral = 93 examples), 
and indicated homogeneity of variance, the standard one-way ANOVA results are reported along with 95\% confidence intervals for effect sizes.

This analysis resulted in statistically significant associations between the multicategory morality ratings and action valence
($F = 198.100, p < 0.001, \eta^{2} = 0.50\,[0.45, 1.00]$), and consequence valence
($F = 218.100, p < 0.001, \eta^{2} = 0.52\,[0.47, 1.00]$). The observed large effect sizes for $\eta^{2}$ indicate that both action and
consequence valence share strong associations with the multicategory morality ratings. Unfortunately, due to the large CIs observed,
the estimation of this effect size is uncertain. These early exploratory results appear promising, however, and it is expected that 
the small number of outliers observed in residual QQ-plots contributed to this estimation uncertainty.

Following the ANOVA analysis, Tukey's honest significant difference (HSD) test was performed to determine pairwise mean differences between groups, the results of which
can be seen in Table \ref{tukey-table}. The results of this analysis show significant differences in valence ratings, for both action and
consequence valence, when comparing the immoral class ratings to that of both discretionary and moral classes. Additionally, Table \ref{tukey-table}
shows that discriminating between the moral and discretionary classes based on valence ratings is difficult. This is evidenced by small mean
differences between these groups' ratings, that in both cases of action and consequence valence did not reach statistical significance at the
$p = 0.05$ level. This corroborates visual clustering-type behaviour observed in Figure \ref{fig:single} for these classes and further confirms
the difficulty of discriminating between them, based on moral valence.

\begin{table}[htbp]
\caption{Tukey's Honest Significant Difference (HSD) Test Results for Action Valence (AV) and Consequence Valence (CV) Differences Between Immoral,
Discretionary, \& Moral Classes}
\begin{center}
\begin{tabular}{|c|c|c|}
\hline 
Groups & Mean Diff. (95\% CI) & p-value \\
\hline
AV Discretionary-Immoral & 0.728 (0.629, 0.826) & $< 0.001$ \\
AV Moral-Immoral & 0.824 (0.709, 0.939) & $< 0.001$ \\
AV Moral-Discretionary & 0.097 (-0.013, 0.207) & $0.097$ \\
CV Discretionary-Immoral & 0.733 (0.638, 0.827) & $< 0.001$ \\
CV Moral-Immoral & 0.831 (0.720, 0.941) & $< 0.001$ \\
CV Moral-Discretionary & 0.098 (-0.007, 0.204) & $0.074$ \\
\hline
\end{tabular}
\label{tukey-table}
\end{center}
\end{table}
Finally, both action and consequence valence had noteworthy positive PCC values with the binary immoral/moral class label, with action valence PCC = 0.703 
($t = -19.717, p < 0.001$) and consequence valence PCC = 0.720 ($t = -20.671, p < 0.001$). These results show that both measurements can be strong
predictors of binary immoral/moral ratings, with consequence valence being a marginally better predictor. Further, these results bolster that of
Table \ref{tab2} where consequence valence was indicated as a higher quality measure (higher $\mu\text{CCC}$) compared to that of action valence.

\subsection{Binary Immoral/Moral Classification using Action and Consequence Valence}
For this task, baseline scores were calculated using the majority class (moral) guess classifier on the training set. This resulted in a zero
information rate model accuracy of 66.50\% along with a MCC of 0. The L2 regularised logistic regression $\lambda$ value learned from
cross-validation was 0.104 and this was used for learning the final logistic regression classifier weights. The baseline, cross-validation, and
test set accuracy \% and MCC obtained from this experimentation are provided in Table \ref{results-table}. It can be seen that very strong
prediction performance was obtained for the simple L2 regularised logistic regression model containing only two input features, where the baseline
score was beaten by a large margin, both in terms of accuracy \%, and more importantly MCC. Absolute improvements of 0.695 (five-fold
cross-validation) and 0.764 (test set) were obtained compared with the majority class baseline classifier MCC. This provides early evidence that
valence can provide effective prediction of binary immoral/moral ratings of text.

\begin{table}[htbp]
\caption{Action and Consequence Valence Binary Immoral/Moral Logistic Regression Classification Results Including Training Majority Class Prediction Only Baseline,
Five-fold Cross-validation Performance (Standard Deviation), and Final Test Set Performance Accuracy \% \& Matthew's Correlation Coefficient (MCC)}
\begin{center}
\begin{tabular}{|c|c|c|}
\hline 
Partition & Accuracy \% & MCC \\
\hline
Train Baseline & 66.50\% & 0 \\
Cross-validation & 86.50\% (5.54\%) & 0.693 (0.134) \\
Test & 89.00\% & 0.764 \\
\hline
\end{tabular}
\label{results-table}
\end{center}
\end{table}

The logistic regression model performed very well for both classes based on valence feature input alone, where high true positive (0.863) and true negative (0.963)
rates were obtained. The most problematic error produced on the test set was where the model predicted that an immoral scenario was moral ($\times 1$). Predicting
immoral text as moral (false positive) facilitates text display to users that has more harm potential than blocking users from seeing moral text (false negative). Hence,
this problematic case was evaluated, both for the action and consequence valence ratings serving as input to the model, along with the text
stimulus that those ratings were based on. The valence input for this case was -0.626 for action valence and -0.566 for consequence valence (EWE gold standard) while
the text input that these ratings were based on was ``telling your friend's business to someone else when you are worried about them, if sharing private information
about a friend with another friend''. This item's ground-truth label was originally \textit{discretionary} but was later collapsed into the moral class. This therefore
indicates annotator error compared with ground-truth, perhaps due to it being a discretionary case (labelled moral for binary prediction). General error analysis was
conducted from 10 test set false negative errors. An interesting result from this analysis was that five out of ten error cases had consequence valence ratings close
to zero. This indicates an opportunity for the development of borderline, hard-to-predict examples in the future, so that models are more capable in reducing these
observed high prevalence errors.

\section{Discussion}\label{Discussion}
The text material presented to annotators from Norm Bank \cite{jiang_investigating_2025} was often morally, and hence emotionally/affectively, evocative.
It has been shown that emotional experience correlates with moral choice in previous work \cite{carmona-perera_valence_2013}. Therefore, it is not
surprising that the valence annotations from this study share significant associations with the Norm Bank moral category labels.
Agreement was additionally found in the results for previously mentioned difficulty in differentiation of discretionary and moral classes \cite{jiang_investigating_2025}.
Differences in means for both action and consequence valence were very small between moral and discretionary class labels and did not reach statistical
significance. Therefore, valence measurements cannot offer much in terms of distinguishing between these moral classes. However, both action and consequence valence were
indicative of immoral vs the other classes. This class was distinguishable from either discretionary or moral classes with large group mean differences observed, all
reaching statistical significance. Further, strong positive correlations between action valence and consequence valence with binary immoral/moral labels were obtained.
From the evaluations undertaken, consequence valence appeared to be more certain across annotators while additionally sharing stronger relationships with moral category
labels. This shows that action valence may be a candidate for removal from the proposed moral valence features due to its inferiority when compared with consequence valence.
Further, for moral valence, it appears slightly more favourable to think of valence in terms of the consequences that follow a valent action/judgement, 
rather than the intrinsic valence of an action/judgement. 

The binary morality classification results obtained on the data using action and consequence valence are strong for the relatively simple modelling approach taken. 
Additionally, emerging evidence from affective computing suggests that frontier LLMs have capabilities of emotion prediction \cite{yongsatianchot_whats_2023}
and cognitive emotional self-appraisal \cite{bhattacharyya_machines_2025}, the simulation of subjective understanding of an LLM's own simulated
emotion. It seems reasonable then, that despite a lack of awareness of this ability unless explicitly prompted, many LLMs may have a capability for,
albeit imperfect, moral estimation of their responses/actions. The results of the present paper suggest, in particular if morality is of primary
concern, that valence inference could be conducted by an AI system on their own generated responses prior to action execution or display to a human.
While valence prediction has traditionally been performed for estimation of a human's valence (this will be illegal without consent under the AI Act),
such an affective-moral computing effort provides another use for this well-studied dimension of human affect. Any implementation of such a feature would 
of course serve as a descriptive moral feature component as part of broader system implementation and should be subject to human oversight \cite{mccormack_trustworthy_2026}. 
Due to the impracticalities of humans monitoring all human-AI interactions, such a system could offer a second-best estimate, compared with human judgement, 
intended to make practically feasible/scalable but not diminish, human compliance workload.

\section{Conclusion and Future Work}\label{Conclusion}
The proposed valence features and experiments presented were intended to discover if subjective valence ratings of moral judgements/actions and
their consequences offer indicative features of morality in text. Based on the experimental results obtained, subjective valence features of text
appear to share strong relationships with both multi-class and binary moral category-rated text. Consequence moral valence was shown 
as the more important of the two proposed valence features. Further, the prediction results obtained were strong, indicating morality estimation
based on moral valence is practicable. While further experimental evaluation of the proposed affective-moral features is required, this work
provides early evidence of the usefulness of moral valence features toward morality estimation of text. 

Some important study limitations are now worth mentioning. The number of annotations provided in this work is small while only one descriptive moral
phenomenon was studied. Therefore, a reasonable first step to advance the present work is to extend the size of the labelled set. 
To this end, further manual annotation could be performed along with data augmentation and weak supervision. The generated descriptive
moral features of the proposed work, while promising, offer an incomplete view of descriptive ethics, perhaps consequentialist-leaning in nature. Hence,
they can, and should, be combined with other descriptive measures such as MFT to offer a more comprehensive characterisation of bottom-up, descriptive ethics.
In addition, for ultimate deployment, a complete implementation of reflective equilibrium with the inclusion of a normative rule-based system should be
pursued while such a system should remain human-aligned and auditable. This leads to further future work including alignment evaluation of human- compared with
LLM-generated moral valence ratings along with multi-theory descriptive \& normative framework implementation and proposals for audit criteria for
effective human oversight. It is hoped that this work advances affective-moral computing efforts that can inform future AI ethics system development.

\bibliographystyle{IEEEtran}
\bibliography{references}

\vspace{12pt}

\end{document}